\documentclass[11pt,a4paper]{article}
\usepackage[hyperref]{emnlp2018}
\usepackage{times}
\usepackage{latexsym}
\usepackage{multirow}
\usepackage{amsmath}
\usepackage{svg}
\usepackage{url}
\usepackage{setspace}
\usepackage{graphicx}
\usepackage{algorithm}
\usepackage[noend]{algpseudocode}
\usepackage{wrapfig}
\usepackage{fp}
\usepackage{CJKutf8}

\aclfinalcopy 

\usepackage{xcolor}

\definecolor{i6blue}{rgb}{0.0, 0.4, 0.8}
\definecolor{i6deepblue}{rgb}{0.0, 0.2, 1.0}
\definecolor{i6green}{rgb}{0.4, 0.8, 0.0}
\definecolor{i6lightblue}{rgb}{0.6, 0.9, 1.0}
\definecolor{i6orange}{rgb}{0.8, 0.4, 0.0}
\definecolor{i6pink}{rgb}{0.8, 0.0, 0.4}

\definecolor{orange}{rgb}{1.0, 0.6, 0.0}
\definecolor{yellow}{rgb}{1.0, 1.0, 0.0}

\definecolor{grey}{rgb}{0.6, 0.6, 0.6}
\definecolor{lightgrey}{rgb}{0.9, 0.9, 0.9}

\definecolor{i6bluedark}{rgb}{0.0156,0.2578,0.5625} 

\def\isechsblue#1    {\textcolor{i6blue} {#1} }
\def\isechsbluedark#1    {\textcolor{i6bluedark} {#1} }
\def\blue#1    {\textcolor{blue} {#1} }
\def\green#1   {\textcolor{green} {#1} }
\def\magenta#1 {\textcolor{magenta} {#1} }

\newcommand\emptyword\epsilon

\newcommand\ngrams{$n$-grams\xspace}

\newcommand\mgrams\ngrams

\newcommand\argmax{\operatorname*{arg max}}

\newcommand{\sourceword}{\ensuremath{f}\xspace}
\newcommand{\targetword}{\ensuremath{e}\xspace}
\newcommand\f\sourceword
\newcommand\e\targetword
\newcommand{\segphrase}{\ensuremath{s}\xspace}
\newcommand{\alignmentword}{\ensuremath{a}\xspace}
\newcommand{\sourcemax}{\ensuremath{J}\xspace}

\newcommand{\segmax}{\ensuremath{K}\xspace}
\newcommand{\sourceindex}{\ensuremath{j}\xspace}

\newcommand{\segindex}{\ensuremath{k}\xspace}

\newcommand\invalignmentword{\ensuremath{B}\xspace}

\renewcommand\j\sourceindex
\renewcommand\k\segindex
\newcommand\J\sourcemax
\newcommand\K\segmax

\newcommand{\sourcephrase}{\ensuremath{\tilde{\sourceword}}\xspace}
\newcommand{\targetphrase}{\ensuremath{\tilde{\targetword}}\xspace}
\newcommand{\sourcesentence}{\ensuremath{\f_1^\J}\xspace}
\newcommand{\targetsentence}{\ensuremath{\e_1^\I}\xspace}
\newcommand{\segsentence}{\ensuremath{\s_1^\K}\xspace}

\newcommand{\alignmentsentence}{\ensuremath{\a_1^{\J}}\xspace}
\newcommand{\invalignmentsentence}{{\ensuremath{\b_0^{\I}}}\xspace}

\newcommand\s\segphrase
\renewcommand\a\alignmentword
\renewcommand\b\invalignmentword
\newcommand\fs\sourcesentence
\newcommand\fp\sourcephrase
\newcommand\es\targetsentence
\newcommand\ep\targetphrase
\newcommand\sse\segsentence
\newcommand\as\alignmentsentence
\newcommand\bs\invalignmentsentence

\newcommand\T{\rule{0pt}{2.2ex}}
\newcommand\pct{\hspace*{-0.4ex}\ensuremath{^{^{[\%]}}}}

\newcommand{\todo}[1]{}
\renewcommand{\todo}[1]{{\color{red} TODO: {#1}}}

\usepackage{tikz}
\usetikzlibrary{shapes,positioning,fit}
\usetikzlibrary{patterns}
\usetikzlibrary{arrows,positioning,arrows.meta}
\usetikzlibrary{decorations}
\usetikzlibrary{fit,calc}
\usetikzlibrary{decorations.pathreplacing}

\title{On The Alignment Problem In Multi-Head Attention-Based Neural Machine Translation}

\author{Tamer Alkhouli, Gabriel Bretschner, and Hermann Ney\\
        Human Language Technology and Pattern Recognition Group \\
        Computer Science Department \\
        RWTH Aachen University \\
        D-52056 Aachen, Germany \\
        {\tt <surname>@i6.informatik.rwth-aachen.de }}

\date{}
\algrenewcommand\algorithmicindent{1.0em}

\begin{document}
\maketitle
\begin{abstract}
This work investigates the alignment problem in state-of-the-art multi-head attention models based on the transformer
architecture. We demonstrate that alignment extraction in transformer models can be improved by augmenting an additional
alignment head to the multi-head source-to-target attention component. This is used to compute sharper  attention weights.
We describe how to use the alignment head to achieve competitive performance. To study the effect of adding the alignment head, we simulate a dictionary-guided
translation task, where  the user wants to guide translation using pre-defined dictionary entries. Using the proposed approach, we achieve up to
$3.8$\% BLEU improvement when using the dictionary, in comparison to $2.4$\% BLEU in the baseline case. We also propose alignment
	pruning to speed up decoding in alignment-based neural machine translation (ANMT), which speeds up translation by a factor of $1.8$ without
loss in translation performance. We carry out experiments on the shared WMT 2016 English$\to$Romanian news task and the
BOLT Chinese$\to$English discussion forum task.

\end{abstract}

\section{Introduction}

\begin{figure*}

	\centering
\includegraphics[scale=0.70]{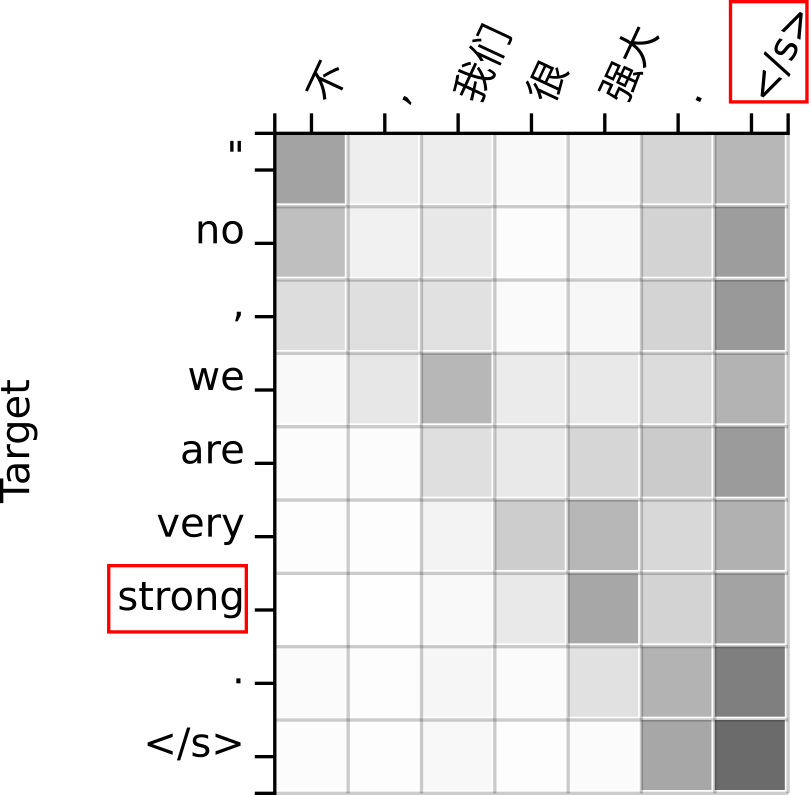}
\includegraphics[scale=0.70]{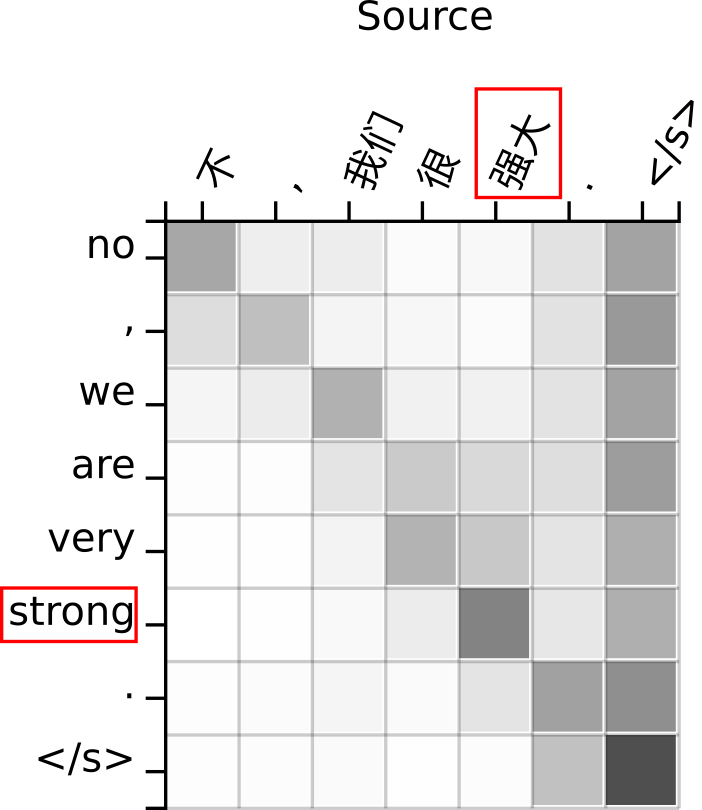}
\includegraphics[scale=0.70]{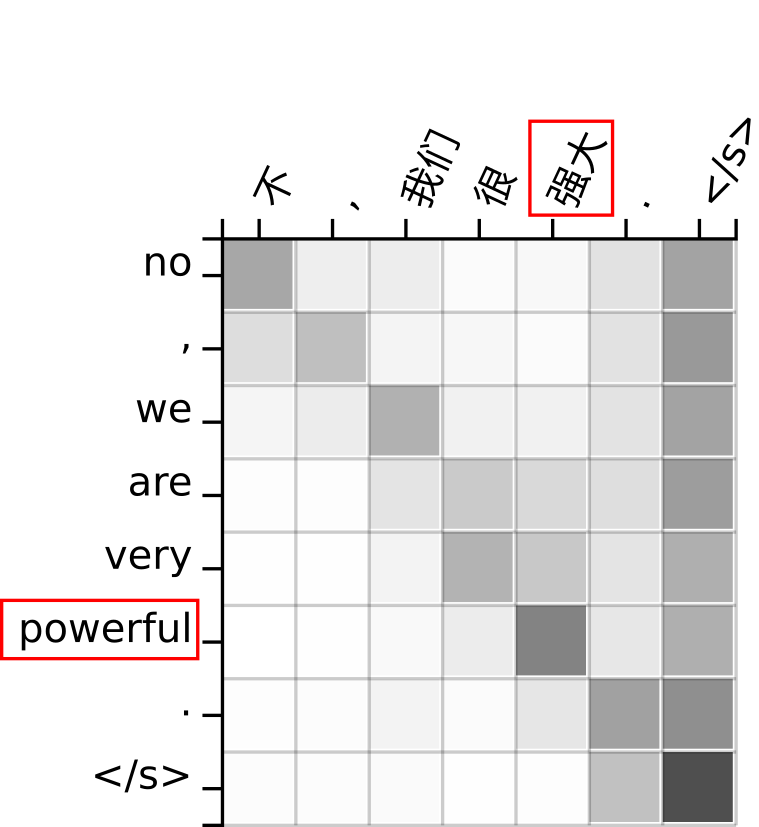}

  \caption{An example from the Chinese$\to$English system. The figures illustrate the accumulated attention weights of the baseline transformer model (left), the
alignment-assisted transformer model (middle), and the alignment-assisted model guided by a dictionary entry. We simulate a scenario where the user wants to
  translate the Chinese word \begin{CJK*}{UTF8}{gbsn}``强大"\end{CJK*}
to ``powerful". Both the baseline and alignment-assisted  transformer models generate the translation ``strong" instead. To enforce the translation, we use the maximum attention weight to determine the source word being translated. Left: The maximum attention of the baseline case incorrectly points to the sentence end when translating the designated Chinese word, therefore we cannot enforce the translation in this case. Middle: The alignment looks sharper because the system has an augmented  alignment head. In this case the maximum attention is
pointing to the correct Chinese word. Right: using the maximum attention, the translation ``strong" is successfully  replaced with the translation ``powerful" as suggested by the user using our proposed alignment-assisted transformer.}
\label{fig:motivation}

\end{figure*}

Attention-based neural machine translation (NMT) \cite{bahdanau14:softAlign} uses an attention layer to determine which part of the input
sequence to focus on during decoding. This component eliminates the need for explicit alignment modeling. In conventional
phrase-based statistical machine translation \cite{koehn2003:pbt}, word alignment is modeled explicitly, making it clear which word or phrase is
being translated. The lack of explicit alignment use in attention-based models makes it harder to determine which
target words are generated using which source words. While this is not necessarily needed for translation itself, alignments can be useful in certain applications, e.g. when the customer wants to enforce specific translation of certain words.

One simple solution is to use maximum attention weights to extract the alignment, but this can result in wrong alignments in
the case where the maximum attention weight is not pointing to the word being translated. Such cases are not uncommon in NMT, making
the use of attention weights as alignment replacement non-trivial
\cite{chatterjee2017:guideddecoding,hasler2018:constraints}. Alignment extraction is even less clear for
transformer models \cite{vaswani2017:transformer}, which currently produce state-of-the-art results. These
models use multiple attention components for each of the stacked decoder layers. In this work we focus our study on these
models since they usually outperform single-attention-head recurrent neural network (RNN) attention models.\footnote{The transformer models won in most of the WMT 2018 news translation tasks: \texttt{http://matrix.statmt.org}.}

Alignment-based NMT \cite{alkhouli16:anmt} uses neural models trained using
explicit hard alignments to generate translation. These systems include explicit alignment modeling, making them more convenient for
tasks where the source-to-target alignment is needed. However, it is not clear whether these systems are able
to compete with strong attention-based NMT systems. \newcite{alkhouli17:alignbiasattention} present results for
alignment-based neural machine translation (ANMT) using
models trained on CPUs, limiting them to small models of 200-node layers, and they only investigate RNN models.
\newcite{wang2018:alignsum} present results using
only one RNN encoder layer, and do not include attention layers in their models. In this work, we investigate the performance
of large and deep state-of-the-art transformer models. We keep  the multi-head attention component and propose to augment it with an additional alignment head, to combine the benefits of the two.
We demonstrate that we can train these models to achieve competitive results in comparison to strong state-of-the-art baselines. Moreover, we demonstrate that this
variant has clear advantage in tasks that require alignments such as  dictionary-guided translation.

Translation in NMT can be performed without explicit alignment. However, there are tasks where
translation needs to be constrained given specific user requirements. Examples include interactive machine translation, and
scenarios where customers demand domain-specific words or phrases to be translated
according to a pre-defined dictionary. We demonstrate that
the explicit use of alignment in ANMT can be leveraged to generate guided translation. Figure~(\ref{fig:motivation})
illustrates an example. The figures are generated using attention weights averaged over all attention components in each system.

The contribution of this work is as follows. First, we propose a method to integrate alignment information into the
 multi-head attention component of the transformer model (Section \ref{sec:lex}). We describe how such models can be trained to maintain the strong baseline
performance while also using external alignment information (Section \ref{sec:train}). We also introduce alignment models that use self-attentive layers for faster evaluation (Section \ref{sec:alignmodel}). Second, we introduce alignment pruning
during search to speed up evaluation without affecting translation quality (Section \ref{sec:prune}). Third, we describe how to extract
alignments from multi-head attention models (Section \ref{sec:align}), and demonstrate that alignment-assisted transformer
systems perform better than baseline systems in dictionary-guided translation tasks (Section \ref{sec:dict}). We present speed and performance results in Section \ref{sec:exp}.

\section{Related Work}
Alignment-based neural models have explicit dependence on the alignment information either at the input or at
the output of the network. They have been extensively and successfully applied on top of
conventional phrase-based systems \cite{sundermeyer14:bidirectional,Tamura14:alignmentRNN,devlin14:nnJoint}. In this
work, we focus on using the models directly to perform standalone neural machine translation.

Alignment-based neural models were proposed in \cite{alkhouli16:anmt} to perform neural machine translation. They mainly used feedforward alignment
and lexical models in decoding. \newcite{alkhouli17:alignbiasattention} used recurrent models instead, and presented an attention component biased using
external alignment information. In this work, we explore the use of transformer models in ANMT instead of recurrent models.

Deriving neural models for translation based on the hidden Markov model (HMM) framework can also be found in
\cite{yang13:alignmentDNN,yu16:noisy}. Alignment-based neural models were also applied to perform
summarization and morphological inflection \cite{yu16:segment}. Their work used a monotonous alignment model,
where training was done by marginalizing over the alignment hidden variables, which is computationally expensive. In
this work,
we use non-monotonous alignment models. In addition, we train using pre-computed Viterbi alignments which
speeds up neural training. In \cite{yu16:noisy}, alignment-based neural models were used to model alignment
and translation from the target to the source side (inverse direction), and a language model was included in
addition. They showed results on a small translation task. In this work, we present results on translation
tasks containing tens of millions of words. We do not include a language model in any of our systems.

There is plenty
 of work on modifying attention models to capture more complex dependencies.
\newcite{cohn2016:alignNMT} introduce structural biases from word-based alignment concepts like fertility and
Markov conditioning. These are internal modifications that leave the model self-contained.
Our modifications introduce alignments as external information to the model. \newcite{arthur16:lexicalNMT} include
lexical probabilities to bias attention. \newcite{chen16:guidedalignment} and \newcite{mi16:supervisedattention} add an extra term dependent on the alignments
to the training objective function to guide neural training. This is only applied during training but not during
decoding. Our work makes use of alignments during training and also during decoding.

There are several approaches to perform constrained translation. One possibility is including this information in training,
but this requires knowing the constraints at training time \cite{crego2016:systran}. Post-processing the hypotheses is
another possibility, but this comes with the downside that offline modification of the hypotheses happens out of context. A
third possibility is to do constrained decoding
\cite{hokamp2017:constraints,chatterjee2017:guideddecoding,hasler2018:constraints,post2018:constrained}. This does not require knowledge of the
constraints at training time, and it also allows dynamic changes of the rest of the hypothesis when the constraints are
activated. We perform experiments where the translation is guided online during decoding. We focus on
the case where translation suggestions are to be used when a word
in the source sentence
matches the source side of a pre-defined dictionary entry. We show that alignment-assisted transformer-based NMT outperforms  standard transformer models
in such a task.

\section{Alignment-Based Neural Machine Translation}
\label{sec:anmt}
Alignment-based NMT divides translation into two steps: (1) alignment and (2) word generation. The system is composed of an alignment
model and a lexical model that can be trained jointly or separately. During translation, the alignment is hypothesized first, and the
lexical score is computed next using the hypothesized alignment  \cite{alkhouli16:anmt}. Hence, each translation hypothesis has an underlying
alignment used to generate it. The alignment model scores the alignment path.

Formally, given a source sentence $f_1^J=f_1...f_j...f_J$, a target sentence $e_1^I=e_1...e_i...e_I$, and an alignment sequence $b_1^I=b_1...b_i...b_I$, where $j=b_i \in \lbrace 1,2,...,J \rbrace$ is the source position aligned to the target position $i \in \lbrace 1,2,...,I \rbrace$, we model translation using an alignment model and a lexical model:

 \begin{align}
 p(e_1^I|f_1^J) = \sum_{b_1^I} &p(e_1^I,b_1^I|f_1^J)  \label{eq:hmmsum} \\
 					   \approx \max_{b_1^I} \prod_{i=1}^I &\underbrace{p(e_i|b_i, b_1^{i-1},e_1^{i-1},f_1^J)}_{\text{lexical model}}
 					   									   \cdot \nonumber \\
&\underbrace{p(b_i|b_1^{i-1},e_1^{i-1},f_1^J)}_{\text{alignment model}}. \nonumber
 \end{align}
Both the lexical model and the alignment model have rich dependencies including the full source context $f_1^J$, the full alignment history $b_1^{i-1}$, and the full target history $e_1^{i-1}$. The lexical model has an extra dependence on the current source position $b_i$.

While previous work focused on RNN structures for the lexical and alignment models \cite{alkhouli17:alignbiasattention}, we
use multi-head self-attentive transformer model structures instead. The next two subsections describe the structural details of
these models.

\subsection{Transformer-Based Lexical Model}
\label{sec:lex}
In this work we propose to use lexical models based on the transformer architecture \cite{vaswani2017:transformer}. This architecture has the following main components:
\begin{itemize}
\item self-attentive layers replacing recurrent layers. These layers are parallelizable due to the lack of sequential dependencies that recurrent layers have.
\item multi-head source-to-target attention: several attention heads are used to attend to the source side. Each attention head computes a normalized probability distribution over the source positions. The attention heads are concatenated. Each decoder layer in the model has its own multi-head attention component.
\end{itemize}

\begin{figure}
        \centering
                \scalebox{0.75}{

\tikzset{
        XOR/.style={
                draw,
                circle,
                append after command={
                [shorten >=\pgflinewidth, shorten <=\pgflinewidth,]
                (\tikzlastnode.north) edge (\tikzlastnode.south)
                (\tikzlastnode.east) edge (\tikzlastnode.west)
        }
    },
    do path picture/.style={%
            path picture={%
                        \pgfpointdiff{\pgfpointanchor{path picture bounding box}{south west}}%
                        {\pgfpointanchor{path picture bounding box}{north east}}%
                        \pgfgetlastxy\x\y%
                        \tikzset{x=\x/2,y=\y/2}%
                        #1
                }
        },
        Wave/.style={
                draw,
                circle,
                do path picture={    
              \draw [line cap=round] (-3/4,0)
              sin (-3/8,1/2) cos (0,0) sin (3/8,-1/2) cos (3/4,0);
              }
        }
}

\tikzset{
        layer/.style={
                draw,
                fill=lightgray,
                rectangle,
                minimum width=1cm, 
                minimum height=2cm,
                rotate=0,
                font=\footnotesize,
                inner sep=0.1 cm,
                outer sep=0.1 cm
        },
        layer_flat/.style={
                draw,
                fill=lightgray,
                rectangle,
                minimum width=2cm, 
                minimum height=1cm,
                rotate=0,
                font=\footnotesize,
                inner sep=0.1 cm,
                outer sep=0.1 cm
        },
}

\begin{tikzpicture}[
	attention/.pic={
		\draw[fill=lightgray](0,3) rectangle (6,4) 
			node[pos=.5] {\small Scaled Dot-Product Attention};
		\foreach \x in {0, 2, 4} 
			\draw[fill=lightgray](\x + 0.5, 1) rectangle (\x+2, 2) 
				node[pos=.5] {\small Linear};
		\foreach \x in {0, 2, 4} 
			\draw[->] (\x + 1.25, 2) -- (\x + 1.25, 3);
	}
	]
	\tikzset{>=Stealth}
	\foreach \x in {0.5,0.25,0.0}\pic[opacity=1-\x] at (\x,\x){attention};	
	\draw [decorate,decoration={brace,amplitude=5pt, mirror}]
		(6.15,3.5) -- (6.75,4.0) node [black,midway,xshift=0.6cm, yshift=-0.1cm] {\footnotesize $K$};

	\node (V) at (1.5,0) {\small $h_1,h_2,...,h_J$};
	\draw[->] (V.north) -- ($(V) + (0,0.9)$);
	\node (K) at (3.5,0) {\small $h_1,h_2,...,h_J$};
	\draw[->] (K.north) -- ($(K) + (0,0.9)$);
	\node (Q) at (5.5,0) {\small $r_{i-1}$};	\foreach \x in {0.5,0.25,0.0}\draw[->,opacity=1-\x] (\x + 3, \x + 4) -- (\x + 3, 5);
	\draw[->] (Q.north) -- ($(Q) + (0,0.9)$);
	\draw[fill=lightgray] (2,5) rectangle (4,6) node[pos=.5] {\small Concat};
	\draw[->] (3,6) -- (3,6.5);
	\draw[fill=lightgray] (2,6.5) rectangle (4,7.5) node[pos=.5] {\small Linear};
	\draw[->] (3,7.5) -- (3,8);

	\node (V2) at (9,0) {\small $h_1,h_2,...,h_J$};
	\draw[->] (V2.north) -- (9, 0.9); 
	\draw[fill=lightgray] (8, 1) rectangle (10,2) node[pos=.5] {\small Alignment};
	\draw[->] (9,2) -- (9,5.5) node[pos=0.1,right]{\small $h_{b_i}$} -- (4, 5.5);

	\end{tikzpicture}}
        \caption{Alignment-assisted multi-head attention component. $h_1,h_2,...,h_J$: the encoder states at all $J$ source positions, $h_{b_i}$: the encoder state at the aligned source position $b_i$, $r_{i-1}$: the previous decoder state, $K$: number of attention heads. Removing the alignment block results in the default multi-head source-to-target attention component of \cite{vaswani2017:transformer}.}
        \label{fig:mhead-attention-alignment-assisted}
\end{figure}
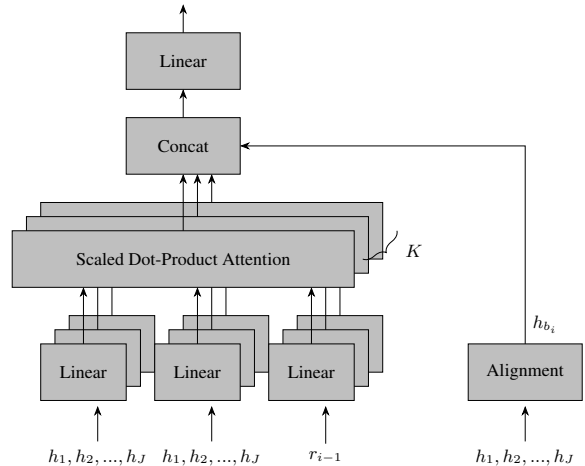

We propose to condition the lexical model on the alignment information. We add a special alignment head
\begin{align*}
\alpha(j|b_{i}) =
     \begin{cases}
       \text{1,} &\quad\text{if}\quad j=b_{i} \\
       \text{0,} &\quad\text{otherwise.} \\
     \end{cases}
\end{align*}
defined for the source positions $j,b_i \in \lbrace 1,2,...,J \rbrace$.
This is a one-hot distribution that has a value of $1$ at position $j$ that matches the aligned position $b_i$. This head is then concatenated to the rest of the attention heads as shown in Figure\ (\ref{fig:mhead-attention-alignment-assisted}).
The one-hot alignment distribution is used similar to attention weights to weight the
encoded source representations, effectively
selecting the representation $h_{b_i}$ which corresponds to the aligned word.

\subsection{Self-Attentive Alignment Model}
\label{sec:alignmodel}

In this work we use self-attentive layers instead of RNN layers in the alignment model. This removes the sequential dependency
of computing RNN activations and allows for parallelization.
We replace the bidirectional RNN encoder of the alignment model by multi-head self-attentive layers as described in
\cite{vaswani2017:transformer}. We also use multi-head self-attentive layers to replace the RNN layers in the decoder part of
the network. There are two main differences when comparing this self-attentive alignment model to the transformer
architecture  described in \cite{vaswani2017:transformer}. (1) The output is a probability distribution over possible source
jumps $\Delta_i=b_i-b_{i-1}$, that is, the model predicts the likelihood of jumping from the previous source position
$b_{i-1}$ to the current source position $b_i$. (2) There is no multi-head source-to-target attention layer as in the
transformer network. Rather, we use a single-head hard attention layer. This layer is not computed like attention weights,
but it is constructed using the previous alignment point $b_{i-1}$ using
\begin{align*}
\alpha(j|b_{i-1}) =
     \begin{cases}
       \text{1,} &\quad\text{if}\quad j=b_{i-1} \\
       \text{0,} &\quad\text{otherwise.} \\
     \end{cases}
\end{align*}
defined for the source positions $j,b_{i-1} \in \lbrace 1,2,...,J \rbrace$.
When multiplied by the source encodings, $\alpha$ effectively selects the source encoding $h_{b_{i-1}}$ of the previous aligned position. This is then summed up with the decoder state $r_{i-1}$.

\subsection{Training}
\label{sec:train}
Our attempts to train the alignment-assisted transformer lexical model from scratch  achieved sub-optimal results.
This could happen because the model could choose to over-rely on the alignment information, risking that the remaining attention
heads would become useless, especially during the early stages of training. To overcome this, we first trained the transformer
baseline parameters without the alignment information until convergence, and used the trained parameters to initialize the alignment-assisted model training. This resulted in better systems compared to training from scratch. We were able to see significant perplexity improvements in the second stage of training indicating that the model was making use of the newly introduced information. Further details are discussed in Section \ref{sec:perf}.

\section{Alignment Pruning}
\label{sec:prune}
Alignment-based decoding requires hypothesizing alignment positions in addition to word translations. The algorithm
is shown in Algorithm (\ref{alg:decoder}). Each lexical hypothesis has an
underlying alignment hypothesis ($activePos$) that is used to compute it (line \ref{alg:decoder:line:batchFFJM}). This is done as a part of beam search. To
speed up decoding, we compute the alignment model output first for all beam entries (line \ref{alg:decoder:line:batchFFAM}). This gives a distribution over the next possible source positions. We prune all
source positions that have a probability below a fixed $threshold$ (lines \ref{alg:decoder:line:beginthreshold}--\ref{alg:decoder:line:endthreshold} ). We only evaluate the lexical model for those positions that survive the threshold.
If the pruning threshold is too aggressive to let any of the source positions survive, pruning is disabled for that  time
step (lines \ref{alg:decoder:line:beginaddall}--\ref{alg:decoder:line:endaddall}).


\definecolor{darkgreen}{rgb}{0.0, 0.2, 0.13}
\definecolor{darkblue}{rgb}{0.0, 0.0, 0.55}

\colorlet{kw}{black}
\colorlet{c}{grey}
\colorlet{v}{black}



\algrenewcommand\algorithmicif{\textbf{\textcolor{kw}{if}}}
\algrenewcommand\algorithmicthen{\textbf{\textcolor{kw}{then}}}
\algrenewcommand\algorithmicelse{\textbf{\textcolor{kw}{else}}}
\algrenewcommand\algorithmicfor{\textbf{\textcolor{kw}{for}}}
\algrenewcommand\algorithmicdo{\textbf{\textcolor{kw}{do}}}
\algrenewcommand\algorithmicforall{\textbf{\textcolor{kw}{for all}}}
\algrenewcommand\algorithmicreturn{\textbf{\textcolor{kw}{return}}}
\renewcommand{\algorithmiccomment}[1]{\textcolor{c}{\hfill$\triangleright${#1}}}

\algnewcommand{\lComment}[1]{\State \textcolor{c}{$\triangleright${#1}}}
\algnewcommand\Continue{\State\textbf{\textcolor{kw}{continue}}}
\algnewcommand\Break{\State\textbf{\textcolor{kw}{break}}}
\algnewcommand\To{\textbf{\textcolor{kw}{ to }}}
\algnewcommand\From{\textbf{\textcolor{kw}{ From }}}
\algnewcommand\Not{\textbf{\textcolor{kw}{ not }}}
\algnewcommand\In{\textbf{\textcolor{kw}{ in }}}
\begin{algorithm}[!t]
\caption{Alignment-Based Pruned Decoding}
\label{alg:JTR}
\linespread{1.1}
\begin{algorithmic}[1]
\small
\Procedure{Translate}{\textcolor{v}{$f_1^J$, $beamSize$, $threshold$}}
\State \textcolor{v}{$hyps$}$\gets$ $initHyp$ \Comment{init. set of partial hypotheses}
\While{\Call{GetBest}{$hyps$} \Not terminated }
\lComment{compute alignment distribution in batch mode}
\State $alignDists \gets$\Call{AlignmentDist}{$hyps$} \label{alg:decoder:line:batchFFAM}
\lComment{hypothesize source alignment points}
\State $activePos \gets \lbrace \rbrace$
\For{$pos$ \From $1$ \To $J$} \label{alg:decoder:line:hypalign}
\lComment{position computed if at least one}
\lComment{beam entry surpasses the threshold}
\For{$b$ \From $1$ \To $beamSize$} \label{alg:decoder:line:hypbeam}
\If{$alignDists[b,pos] > threshold$} \label{alg:decoder:line:beginthreshold}
\State $activePos.Append(pos)$
\Break \label{alg:decoder:line:endthreshold}
\EndIf
\EndFor
\EndFor
\lComment{evaluate all positions if none survived pruning}
\If{$activePos$ is empty}\label{alg:decoder:line:beginaddall}
\State $activePos \gets \lbrace 1,...J \rbrace$
\EndIf \label{alg:decoder:line:endaddall}
\lComment{compute lexical distributions of all}
\lComment{hypotheses in hyps in batch mode}
\State $lexDists \gets$ \Call{LexicalDist}{$hyps$, $activePos$}\label{alg:decoder:line:batchFFJM}
\lComment{combine lexical and alignment scores}
\State $hyps \gets Combine(lexDists,alignDists)$
\lComment{prune to fit the beam}
\State $hyps \gets Prune(hyps,beamSize)$
\EndWhile
\lComment{return the best scoring hypothesis}
\State \Return \textcolor{v}{\Call{GetBest}{$hyps$} }
\EndProcedure
\end{algorithmic}

\label{alg:decoder}
\end{algorithm}

\begin{table*}[!t]
\begin{center}
\begin{tabular}{|l|rr|rr|} 
\hline
        		   							& \multicolumn{2}{c|}{WMT 2016}		& \multicolumn{2}{c|}{BOLT} \T\\
     	      							& \bf English	& \bf Romanian	& \bf Chinese & \bf English \T\\
\hline
{\tt Train} sentence pairs 				& \multicolumn{2}{c|}{604K} 	& \multicolumn{2}{c|}{4.1M} \T\\
{\tt Train} running  words					& 15.5M & 15.8M				& 80M   & 88M \\
{\tt Dev} sentence pairs 				& \multicolumn{2}{c|}{1000} 	& \multicolumn{2}{c|}{1845} \T\\
{\tt Test} sentence pairs 				& \multicolumn{2}{c|}{1999} 	& \multicolumn{2}{c|}{1124} \T\\
Vocabulary 	    							& 92K 	& 128K		& 380K  	& 815K \\
Neural network vocabulary 				& 50K 	& 50K				& 50K  	& 50K \\
\hline
\end{tabular}
\caption{Corpora statistics.}
\label{tab:tasks}
\end{center}
\end{table*}
\begin{table*}[!t]

\centering
\setlength\tabcolsep{2pt}%

        \begin{tabular}{|c|l|c|c|cc|c|cc|}
        \hline
            \multicolumn{3}{|c|}{}		 			 & \multicolumn{3}{c|}{WMT En$\to$Ro}& \multicolumn{3}{c|}{BOLT Zh$\to$En} \T\\
            \multicolumn{3}{|c|}{}		 			 & \multicolumn{3}{c|}{newstest2016} & \multicolumn{3}{c|}{test}   \T\\               \hline
        \# & System	 	  					& Layer size & PPL & BLEU \pct & TER \pct & PPL & BLEU \pct & TER \pct  \\
        \hline
 \multicolumn{9}{|c|}{baselines} \\ \hline
1& Attention baseline 					& 1000 	& 10.2   & 24.7 & 58.9 & 8.0 & 20.0 &  65.6 \T \\
2& Transformer baseline 					& 2048 	& \hphantom06.2 & 27.9 & 54.6 & 6.0 & 22.5 & 62.1 \T \\
3& \cite{alkhouli17:alignbiasattention} 	& 200	& -   & 24.8 & 58.1 & - & -   &	 - \T \\
\hline
 \multicolumn{9}{|c|}{this work} \\ \hline
4& RNN Attention align.-biased    		& 1000	& 7.2   & 26.4 & 56.1 & 5.6 &  19.6 & 62.3 \T \\
		5& Align.-assisted Transformer    		& 2048	& \textbf{5.0} & \textbf{28.1} & \textbf{54.3} & \textbf{4.7} & \textbf{22.7} & \textbf{61.8} \T \\
\hline
        \end{tabular}
        \caption{Translation results for the WMT 2016 English$\to$Romanian task and the BOLT Chinese$\to$English task. We include the lexical model perplexities.}
        \label{tab:baselinecomparison}

\end{table*}

\section{Alignment Extraction}
\label{sec:align}

We use attention
weights to extract the alignments  at each time step during
decoding. We look up the source word having the maximum accumulated attention weight
\begin{align*}
j(i) = \argmax_{\hat{j} \in \lbrace 1 ... J \rbrace } \left\lbrace \sum_{l=1}^L  \sum_{k=1}^K  \alpha_{i,k,l}(\hat{j}) \right\rbrace
\end{align*}
where $K$ is the number of attention heads per decoder layer, $L$ is the number of decoder layers, $\alpha_{i,k,l}(\hat{j})$
is the attention weight at source position $\hat{j} \in \lbrace 1, ..., J \rbrace$ for target position $i$ of the $k$-th head computed for the the $l$-th decoder
layer. This is an extension of using maximum attention weights in single-head attention models \cite{chatterjee2017:guideddecoding}. In the alignment-assisted transformer, the aligned position is
given by:
\begin{align*}
j(i,j') \hspace{-0.5mm} =  \hspace{-0.5mm} \argmax_{\hat{j} \in \lbrace 1 ... J \rbrace } \left\lbrace  \sum_{l=1}^L \hspace{-1mm} \bigg( \sum_{k=1}^K \alpha_{i,k,l}(\hat{j}) + \alpha(\hat{j}|j')  \hspace{-1mm}  \bigg)  \hspace{-1mm}  \right\rbrace
\end{align*}
where $j' \in \lbrace 1,...,J \rbrace$ is the hypothesized source position during search, and $\alpha(\hat{j}|j')$ is the
alignment indicator which is equal to $1$ if $\hat{j}=j'$ and zero otherwise. This effectively gives a preference for the
hypothesized position over all other positions. Note that the hypothesized positions are scored during translation using the
alignment model described in Section \ref{sec:alignmodel}.

\section{Experiments}
\label{sec:exp}
We run experiments on the WMT 2016
English$\to$Romanian news task,\footnote{\texttt{http://www.statmt.org/wmt16/}}
and on BOLT Chinese$\to$English which is a
discussion forum task. The corpora statistics
are shown in Table~(\ref{tab:tasks}).

All transformer models use $6$ encoder and $6$ decoder self-attentive layers. We use $8$ scaled dot product attention heads and augment an
additional alignment head to the source-to-target attention component. We use an embedding size of $512$. The size of feedforward layers is $2048$ nodes. We use source and target weight tying for the WMT English$\to$Romanian task, and no tying for BOLT Chinese$\to$English.

The structure of the RNN models is as follows. The English$\to$Romanian lexical and alignment models use $1$ bidirectional encoder layer.
The Chinese$\to$English models have
$1$ bidirectional encoder and $3$ stacked unidirectional encoder layers. All models use $2$ decoder layers. The baseline attention models have
similar structures.  We use LSTM layers of $1000$ nodes and
embeddings of size $620$. We train using the Adam optimizer \cite{kingma2015:adam}. All alignment models predict source jumps of maximum width of $100$ source positions (forward and backward).

The alignments used during training are the
result of IBM1/HMM/IBM4 training using GIZA++ \citep{och03:align}.
All results are measured in case-insensitive BLEU[\%] \cite{papineni02:bleu}.
TER[\%] scores are computed with \emph{TERCom}  \cite{snover06:ter}.
We implement the models in Sockeye \cite{hieber2017:sockeye}, which allows efficient training of large models on GPUs.

\subsection{Performance Comparison}
\label{sec:perf}
\begin{table*}[!t]
\centering
\setlength\tabcolsep{2pt}%

        \begin{tabular}{|c|l|ccc|ccc|}
        \hline
            \multicolumn{2}{|c|}{}		 			 & \multicolumn{3}{c|}{WMT En$\to$Ro} & \multicolumn{3}{c|}{BOLT Zh$\to$En}  \T\\
        \# & Alignment 	 &  \#entries \hspace{1mm} 	& BLEU \pct & TER \pct &  \#entries  \hspace{1mm}  & BLEU \pct & TER \pct \\
        \hline
1& Transformer baseline	 				& - & 27.3 & 55.6 & - & 24.2 & 61.5 \T \\
2& + dictionary 							& 3.1K & 29.7 & 55.4 & 4.6K& 25.5 & 61.0 \T \\
3& Alignment-assisted Transformer  		& - & 27.2 & 55.5 & - & 24.2 & 60.8 \T \\
4& + dictionary							& 3.1K & \textbf{31.0} & \textbf{53.0} & 4.6K & \textbf{26.4} & \textbf{58.6} \T \\
\hline
        \end{tabular}
        \caption[]{Improvements after using the dictionary of the development sets. The tokenized references of the English$\to$Romanian and Chinese$\to$English development sets have $26.7$K and $46.6$K running words respectively. }
        \label{tab:dictionary}

\end{table*}
Table~(\ref{tab:baselinecomparison}) presents results on the two tasks.  The RNN attention (row~1) and transformer (row~2)
baselines are shown. The transformer baseline outperforms the attention baseline by a large margin.  We also include
the English$\to$Romanian system of \newcite{alkhouli17:alignbiasattention}. This is an alignment-based RNN attention system
which uses 200-node layers. We also trained our own alignment-based RNN attention system using larger layers of $1000$ nodes.
This is shown in row~4. Our RNN system outperforms the previously published alignment-based results (row~3) by $1.6$\% BLEU and $2.0$\% TER. This is due to the increase in model size.

Our proposed alignment-assisted transformer system is shown in row~5. This system outperforms the RNN alignment-based system of row~4 by $1.7$\% BLEU on the English$\to$Romanian task, establishing a new state-of-the-art result for alignment-based neural machine translation. We also achieve $3.1$\% BLEU improvement over our
RNN alignment-biased attention system on the Chinese$\to$English task. In comparison to the transformer baseline (row~2), the proposed system achieves similar performance on both tasks. We compare the development perplexity to check whether the lexical model makes use of the alignment information. Indeed, the baseline transformer development perplexity drops from $6.2$ to $5.0$ on English$\to$Romanian and from $6.0$ to $4.7$ on Chinese$\to$English, indicating that the model is making use of the alignment information.

\subsection{Decoding Speed Up}
\begin{figure}

\small
\hspace{-2.5mm} \input{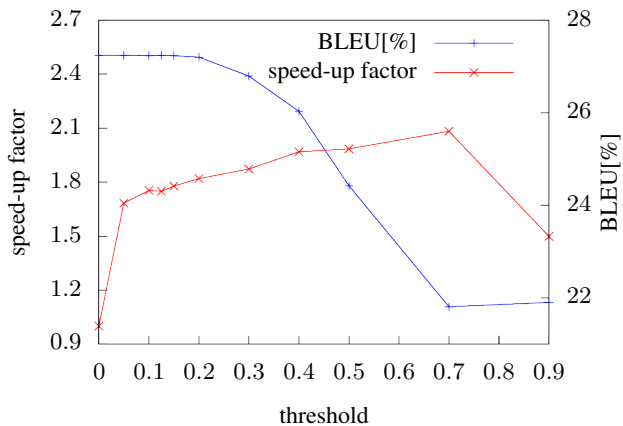}
	\caption[]{Speed up and translation quality in BLEU vs. pruning threshold on the WMT English$\to$Romanian task.}
\label{fig:speedup}

\end{figure}
Figure~(\ref{fig:speedup})
shows the speed-up factor and performance in BLEU over different threshold values. The speed-up factor is computed against the  no-pruning case (i.e. threshold $0$). The batch size used in these experiments is $5$. We speed up translation by a factor of $1.8$
without loss in translation quality at threshold 0.15. Higher threshold values result in more aggressive pruning and hence a degradation in translation quality. It is
interesting to note that at threshold 0.05 we achieve a speed up of $1.7$, implying that
significant pruning happens at low threshold values. At high threshold values, speed starts to go down, since we have more cases where no alignment points survive the threshold, in
which case pruning is disabled as discussed in Algorithm (\ref{alg:decoder}, lines \ref{alg:decoder:line:beginaddall}--\ref{alg:decoder:line:endaddall}).

\section{Dictionary Suggestions}
\label{sec:dict}
We evaluate the use of attention weights as alignments in a dictionary suggestion task, where a pre-defined dictionary of suggested
one-to-one translations is given. We perform a relaxed form of constrained translation, i.e. we do not ensure that the suggestion will
make it to the translation. To this end, we use attention
weights to extract the alignments  at each time step during
decoding as described in Section \ref{sec:align}. We look up the source word $f_{j(i)}$ having the maximum accumulated attention weight in the dictionary. If the word  matches the source-side of a dictionary entry, we enforce the translation to match the dictionary suggestion $e(f_{j(i)})$ by setting an infinite cost for
all but the suggested word.

We create a simulated dictionary using the reference side of the development set. We map the reference to the source words
using IBM4 alignment. The development set is concatenated with the training data to obtain
good-quality alignment. We exclude English stop words,\footnote{Long stop list: \texttt{https://www.ranks.nl/stopwords}} and only use source words aligned one-to-one to target words. We
include up to $4$ dictionary entries per sentence, and add reference translations only if they are not part of the baseline
(i.e. unconstrained) translation, similar to \cite{hasler2018:constraints}.

Table~(\ref{tab:dictionary}) shows results for the dictionary suggestions task described in Section (\ref{sec:dict}). The English$\to$Romanian dictionary covers $11.6$\% of the reference set, while the Chinese$\to$English dictionary has $9.9$\% coverage. We observe larger improvement when using the dictionary entries in the alignment-assisted transformer system in comparison to the transformer baseline systems. Our system improves BLEU by $3.8$\%, while the baseline is improved only by $2.4$\% BLEU on the English$\to$Romanian task. We also observe larger improvements in the Chinese$\to$English case. This suggests that the maximum attention weights in alignment-assisted systems can point more accurately to the word being translated, allowing the use of more dictionary entries. As shown in Figure~(\ref{fig:motivation}), the accumulated attention weights are sharper when the system has an augmented alignment head. This explains the larger improvements our systems achieve.

\section{Conclusion }
We proposed augmenting transformer models with an alignment head to help extract alignments in scenarios such as dictionary-guided translation. We demonstrated that the alignment-assisted systems can achieve competitive performance compared to strong
transformer baselines. We also showed that the alignment-assisted systems outperformed standard transformer models when used for dictionary-guided translation on
two tasks. Finally, we achieved a speed-up factor of $1.8$ by pruning alignment hypotheses in alignment-based decoding while maintaining translation quality. In future work we plan
to investigate alternative pruning methods like histogram pruning. We also plan to investigate the performance of alignment-assisted transformer models in constrained decoding settings,
where the user demands specific translation of certain words.

\section*{Acknowledgments}
\begingroup
\setlength{\columnsep}{5pt}%
\setlength{\intextsep}{0pt}%
\includegraphics[width=0.45\textwidth]{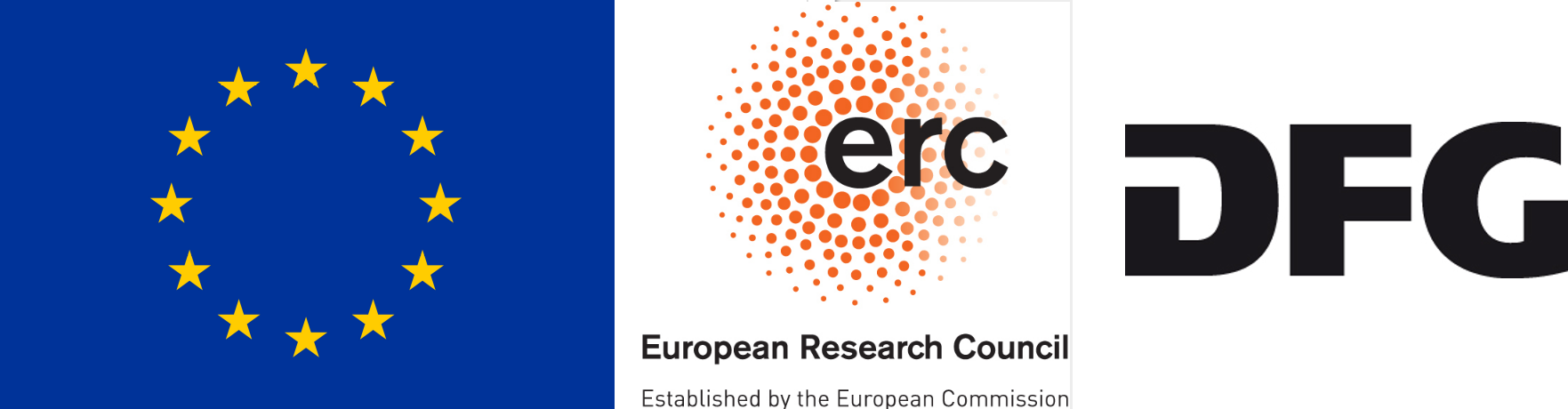}

\noindent
This research
has received funding from the European Research Council (ERC)
(under the European Union's Horizon 2020 research and innovation
programme, grant agreement No 694537, project ``SEQCLAS") and
the Deutsche Forschungsgemeinschaft (DFG; grant agreement NE 572/8-1,
project ``CoreTec"). Tamer Alkhouli was partly funded by the 2016 Google PhD fellowship for North America,
Europe and the Middle East.
The work reflects only the authors' views and none of the funding parties
is responsible for any use that may be made of the information it contains.

 \bibliography{translation}
 \bibliographystyle{acl_natbib_nourl}

 \appendix

\end{document}